\documentclass{article}

\usepackage{microtype}
\usepackage{graphicx}
\usepackage{subcaption}
\usepackage{booktabs} 
\usepackage{float}
\usepackage{placeins}

\usepackage{hyperref}

\usepackage[preprint]{icml2026}

\usepackage{amsmath}
\usepackage{amssymb}
\usepackage{mathtools}
\usepackage{amsthm}

\usepackage[capitalize,noabbrev]{cleveref}

\theoremstyle{plain}
\newtheorem{theorem}{Theorem}[section]
\newtheorem{proposition}[theorem]{Proposition}

\theoremstyle{definition}

\theoremstyle{remark}

\hypersetup{
    pdftitle={Model Order Selection for Overparameterized Hidden Markov Models},
    pdfauthor={Ghaleb Khalil and Tim Leung},
    pdfsubject={Research Preprint},
    pdfkeywords={Hidden Markov Models, Model-Order Selection, Overparameterization, Behavioral Similarity, Optimal Transport}
}

\begin{document}
\twocolumn[
  \icmltitle{Model Order Selection for Overparameterized Hidden Markov Models}



  \icmlsetsymbol{equal}{*}

  \begin{icmlauthorlist}
    \icmlauthor{Ghaleb Khalil}{uw}
    \icmlauthor{Tim Leung}{uw}
  \end{icmlauthorlist}

  \icmlaffiliation{uw}{Department of Applied Mathematics, University of Washington, Seattle, Washington, USA}
  \icmlcorrespondingauthor{Ghaleb Khalil}{ghaleb@uw.edu}

  \icmlkeywords{Hidden Markov Models, Model-Order Selection, Overparameterization, Behavioral Similarity, Optimal Transport, Limit Order Books}

  \vskip 0.3in
]



\printAffiliationsAndNotice{}  

\begin{abstract}
In an overparameterized HMM, additional fitted states may be weakly occupied, behaviorally redundant, or unstable rather than representing new dynamics. Therefore, selecting the optimal number of hidden states requires distinguishing nominal states from genuinely distinct latent structure. We propose a novel method called the Behavioral-Complexity Model-Order Selection (BCMOS). It combines posterior state occupancy with a similarity measure that incorporates both emission and transition behaviors. The method yields an effective behavioral state count \(K_{\mathrm{BEH}}\) that is evaluated across candidate orders $K$ to form an effective-complexity curve and converted into a nominal model-order estimate using segmented regression and stability of the estimated breakpoint across a grid of behavioral weights. The diagnostic experiments and theoretical analysis identify scalar compression as an important limitation of the current procedure. Applying this methodology to the MSFT limit order book dataset, BCMOS selects a parsimonious interior order \(K=15\), while BIC attains its minimum at the upper candidate boundary \(K=30\). Overall, behavioral complexity provides a complementary perspective on HMM order selection by focusing on the emergence of distinct latent structure rather than likelihood improvement alone.
\end{abstract}

\section{Introduction}

The model order, denoted by $K$, is the number of hidden states. Its magnitude affects the parameter estimation and model interpretation. We consider the HMM model-order selection problem. The nominal number of states $K$ can be difficult to interpret, typically due to underfitting or overfitting. When $K$ is small, distinct latent regimes are potentially merged together, whereas a large $K$ may lead to extra fitted states and nonidentifiability issues of the HMM parameters \citep{gassiat2014}.

\subsection{Existing Approaches to HMM Model-Order Selection}

In the literature, several methods have been proposed to choose the number of hidden states $K$ in an HMM. The classical penalized-likelihood methods are AIC and BIC, which add penalties to discourage unnecessary complexity as $K$ grows, so their selection balances fit against model complexity \citep{akaike1974,schwarz1978}. Bayesian approaches introduce prior distributions on model parameters, and some formulations also place a prior on the model order $K$. Under suitable prior assumptions, consistency results for Bayesian HMM order estimation have been established \citep{gassiat2014}. More recently, Chen et al.\ proposed using marginal likelihood for HMM order selection by integrating over the parameter space. Their method includes consistency results and a computational procedure for estimating the marginal likelihood \citep{chen2024}.

The problem remains difficult for overfitted Gaussian HMMs because the parameters can become non-identifiable. Therefore, these issues make the usual regularity conditions behind likelihood-based model-selection theory harder to satisfy \citep{gassiat2014,chen2024}.

These existing approaches are not specifically constructed to examine whether each additional fitted state contributes genuinely distinct latent structure; instead, they score the candidate model as a whole.

\subsection{Overparameterization and the Model-Order Gap}

When the fitted HMM has more nominal states than are needed to represent the underlying process, these extra states do not necessarily correspond to genuinely new latent regimes \citep{gassiat2014}. Therefore, overparameterization can appear through weak occupancy, when some extra states receive very little posterior support; fragmentation, when posterior mass associated with one meaningful regime is spread across several fitted states; redundancy, when multiple fitted states represent very similar behavior, have similar emission distributions, and may also have similar transition behavior \citep{rousseau2011,gassiat2014}; and finally state-label non-identifiability, when permuting state labels can represent the same observable process, so the numerical labels assigned to hidden states have no intrinsic meaning \citep{cappe2005}. As a consequence, the nominal number of hidden states $K$ does not reflect the true complexity represented by the fitted HMM, a large-$K$ model can concentrate most posterior probability on a much smaller subset of states, and several nominal states may represent structure that is already captured elsewhere in the model.

In contrast to the traditional thinking of asking whether adding another state improves model fit, we ask whether adding another state actually introduces new latent structure. Can the true model order be inferred from how much genuinely new latent structure appears as the candidate order $K$ increases?

\section{Background and Problem Formulation}

\subsection{HMM Model-Order Problem}

Let the observed sequence be
$
Y_{1:T}=(Y_1,\ldots,Y_T), \qquad Y_t\in\mathbb{R}^d,
$
with $T$ being the sequence length and $d$ being the dimension of each observation. The hidden-state process is
$
Z_{1:T}=(Z_1,\ldots,Z_T),
$
with each hidden state satisfying
$
Z_t\in\{1,\ldots,K\},
$
$K$ being the nominal number of hidden states. $Z_t$ is a first-order, time-homogeneous Markov chain, with
$
\pi_k=\Pr(Z_1=k)
$
being the initial distribution and
$
A_{ij}=\Pr(Z_{t+1}=j\mid Z_t=i)
$
being the transition matrix. As for the constraints, the probability vector $\pi$ satisfies $\pi_k \geq 0$ and $\sum_k \pi_k = 1$, and each row $A_i$ of $A$ is a probability vector satisfying $A_{ij} \geq 0$ and $\sum_j A_{ij} = 1$.

For the observation model, we use the Gaussian emission model conditional on the hidden state $Z_t=k$:
$
Y_t\mid Z_t=k\sim\mathcal{N}(\mu_k,\Sigma_k),
$
where $\mu_k$ is the state-specific mean and $\Sigma_k$ is the state-specific covariance matrix satisfying $\Sigma_k \succ 0$. Therefore, for a $K$-state HMM,
$
\theta_K=
\left(
\pi,A,\{\mu_k,\Sigma_k\}_{k=1}^{K}
\right)
$
is the model parameter \citep{rabiner1989}.

Let $K_0$ denote the minimal number of hidden states required to represent the data-generating distribution within the assumed Gaussian HMM family,
$\mathcal{K}=\{K_{\min},\ldots,K_{\max}\}$
the candidate orders, and
$
\widehat K=\widehat K(Y_{1:T})
$
the estimator.
The objective is for $\widehat K$ to recover $K_0$. For every $K\in\mathcal{K}$, we estimate $\theta_K$ using the Baum--Welch algorithm as a local optimization procedure and retain the successful restart with the largest final observed-data log-likelihood $\ell_K(\theta)=\log p_\theta(Y_{1:T})$ \citep{cappe2005}. Therefore, for each candidate $K$, the procedure produces a fitted model
$
\widehat{\theta}_K.
$

\subsection{Posterior Occupancy and Occupancy-Based Effective Count}

For each candidate $K$, assuming that we already have the fitted parameter estimate $\widehat{\theta}_K$, we define the smoothed posterior probability that the hidden state at time $t$ equals $k$ as
$
\gamma_t(k)
=
\Pr\!\left(Z_t=k\mid Y_{1:T},\widehat{\theta}_K\right)
$
\citep{cappe2005}.
Moreover, we define the posterior occupancy as the average of the posterior probabilities over time:
$
p_k
=
\frac{1}{T}\sum_{t=1}^{T}\gamma_t(k),
$
with the occupancies satisfying
$
\sum_{k=1}^{K}p_k=1.
$
Thus, $p=(p_1,\ldots,p_K)^\top$ is a probability vector over the fitted hidden states.
$p_k$ measures how much posterior support state $k$ receives across the full observed sequence. A very small $p_k$ means that the fitted state is weakly supported by the data, and this motivates assigning weakly supported states less weight in effective latent complexity. In a normal HMM fit, the raw nominal $K$ does not account for this because every state is counted equally. Therefore, we introduce the occupancy-based effective count
\[
K_{\mathrm{eff}}
=
\frac{1}{\sum_{k=1}^{K}p_k^2},
\]
defined as the inverse-Simpson effective count applied to posterior occupancies \citep{hill1973,leinster2012}. 
Since $p$ is a probability vector,
$
\frac{1}{K}
\leq
\sum_{k=1}^{K}p_k^2
\leq
1,
$
and therefore
$
1
\leq
K_{\mathrm{eff}}
\leq
K.
$
If
$
p_k=\frac{1}{K}\qquad \forall k,
$
then $K_{\mathrm{eff}}=K$, and in this case, all nominal states receive equal posterior support. However, if $K_{\mathrm{eff}}<K$, then the posterior mass is concentrated on only a subset of the states, so the more uneven the occupancies, the smaller $K_{\mathrm{eff}}$. Therefore, $K_{\mathrm{eff}}$ is the number of equally occupied states that would produce the same degree of posterior concentration; it adjusts nominal $K$ for unequal state support.

\subsection{Behavioral Similarity Between Hidden States}

Posterior occupancy, by definition, does not tell us whether two fitted states represent genuinely different latent behavior; therefore, state similarity must also be considered. A hidden state is characterized by the distribution of observations it emits and the way it transitions to other states, so similarity should incorporate both emission behavior and transition behavior.

To measure how different the two observation emission distributions are, we use the normalized Hellinger distance convention between their Gaussian emission distributions \citep{gibbs2002}:
\[
H^2(P,Q)
=
1-\int \sqrt{p(y)q(y)}\,dy,
\]
so that
$
H(P,Q)\in[0,1].
$

For states $i$ and $j$,
\[
D^{\mathrm{em}}_{ij}
=
H\!\left(
\mathcal N(\mu_i,\Sigma_i),
\mathcal N(\mu_j,\Sigma_j)
\right)
\]
is suitable for a pairwise state-distance matrix, as it is symmetric:
$
D^{\mathrm{em}}_{ij}=D^{\mathrm{em}}_{ji},
$
equal to zero when the emission distributions are identical, and it lies in $[0,1]$. To measure transition dissimilarity, we compare the transition rows $A_i$ and $A_j$ by using the discrete optimal transport Wasserstein distance \citep{peyre2019}:
\[
W_D(A_i,A_j)
=
\min_{\Gamma\in\Pi(A_i,A_j)}
\sum_{\ell=1}^{K}\sum_{m=1}^{K}
\Gamma_{\ell m}D_{\ell m}.
\]
With the symmetric zero-diagonal ground cost matrix
$
D\in[0,1]^{K\times K},
$
the set of admissible transport couplings
$
\Pi(A_i,A_j)
=
\left\{
\Gamma\in\mathbb R_+^{K\times K}:
\Gamma\mathbf 1=A_i,\;
\Gamma^\top\mathbf 1=A_j
\right\},
$
$\Gamma_{\ell m}$ is the amount of transition probability mass transported from destination state $\ell$ to destination state $m$, and $D_{\ell m}$ is the cost of matching destination state $\ell$ with destination state $m$. Optimal Transport is preferable for comparing transition rows, as the hidden state labels have no intrinsic meaning, so a direct norm
$
\|A_i-A_j\|_1
$
treats the different coordinates as different states regardless of their behavioral relationship; therefore, two transition rows may place mass on different numerical destination labels even when those destination states are behaviorally similar. Moreover, Optimal Transport creates a recursive problem, as it allows a small transition distance when probability mass only needs to move between behaviorally similar destination states; then, the transition distance between $i$ and $j$ depends on the behavioral distances among their possible destination states \citep{ferns2004}.

The behavioral-distance operator is
\[
\mathcal T_\alpha(D)_{ij}
=
\alpha D^{\mathrm{em}}_{ij}
+
(1-\alpha)W_D(A_i,A_j)
\]
for $\alpha\in(0,1]$; therefore,
$
D^{\mathrm{BEH}}
=
\mathcal T_\alpha(D^{\mathrm{BEH}})
$
and
\[
D^{\mathrm{BEH}}_{ij}
=
\alpha D^{\mathrm{em}}_{ij}
+
(1-\alpha)
W_{D^{\mathrm{BEH}}}(A_i,A_j).
\]
With $\alpha=1$, the behavioral distance depends only on the emissions, and with a smaller $\alpha$, the weight is placed more on transition behavior. This recursive-distance construction is a specialization of
discounted Kantorovich fixed-point metrics for behavioral similarity
in Markov processes and Markov decision processes
\citep{ferns2004,ferns2011}. The resulting behavioral distance is
well defined as follows.

\begin{proposition}[Well-posedness of the behavioral distance]
\label{prop:behavioral_distance}
Fix a fitted $K$-state HMM and $\alpha\in(0,1]$. Let
$\mathcal{D}_K$ denote the set of $[0,1]$-valued pseudometrics on
$\{1,\ldots,K\}$. Then $\mathcal T_\alpha$ maps
$\mathcal{D}_K$ into itself and, for all
$D,D'\in\mathcal{D}_K$,
\[
\|\mathcal T_\alpha(D)-\mathcal T_\alpha(D')\|_\infty
\leq
(1-\alpha)\|D-D'\|_\infty.
\]
Consequently, $\mathcal T_\alpha$ has a unique fixed point
$D^{\mathrm{BEH}}\in\mathcal{D}_K$. Moreover, for any
$D^{(0)}\in\mathcal{D}_K$, the iteration
\[
D^{(r+1)}=\mathcal T_\alpha(D^{(r)})
\]
converges to $D^{\mathrm{BEH}}$, with
\[
\|D^{(r)}-D^{\mathrm{BEH}}\|_\infty
\leq
(1-\alpha)^r
\|D^{(0)}-D^{\mathrm{BEH}}\|_\infty.
\]
\end{proposition}

A proof is provided in Appendix~\ref{app:proofs}.

Furthermore, we convert the behavioral distance into a behavioral
similarity:
\[
S^{\mathrm{BEH}}_{ij}
=
1-D^{\mathrm{BEH}}_{ij},
\]
with
$
0\leq S^{\mathrm{BEH}}_{ij}\leq1,
\qquad
S^{\mathrm{BEH}}_{ii}=1.
$

A large value of $S^{\mathrm{BEH}}_{ij}$ indicates that states $i$
and $j$ have similar emission distributions and transition behavior.

\subsection{Behavioral Effective Complexity}

To account for posterior occupancy and behavioral similarity, we adopt the similarity-sensitive diversity framework of Leinster and Cobbold \citep{leinster2012}. Using the posterior occupancy vector $p$ defined in Section 2.2 and the behavioral similarity matrix $S^{\mathrm{BEH}}$ defined in Section 2.3, the behavioral effective complexity is defined as
\[
K_{\mathrm{BEH}}
=
\frac{1}{p^\top S^{\mathrm{BEH}}p},
\]
using the order-$2$ form of the Leinster--Cobbold diversity measure. Expanding the denominator gives
\[
p^\top S^{\mathrm{BEH}}p
=
\sum_{i=1}^{K}p_i^2
+
2\sum_{i<j}p_i p_j S^{\mathrm{BEH}}_{ij}.
\]
The first term,
$
\sum_{i=1}^{K}p_i^2,
$
represents the occupancy concentration already captured by $K_{\mathrm{eff}}$, while the second term adjusts this concentration according to pairwise behavioral similarity. Increasing $S^{\mathrm{BEH}}_{ij}$ increases the denominator and therefore decreases the effective contribution of states $i$ and $j$. 
Moreover, the denominator satisfies
\[
\sum_{i=1}^{K}p_i^2
\leq
p^\top S^{\mathrm{BEH}}p
\leq
\left(\sum_{i=1}^{K}p_i\right)^2
=
1,
\]
since
$
0\leq S^{\mathrm{BEH}}_{ij}\leq1
$
and
$
S^{\mathrm{BEH}}_{ii}=1.
$

Therefore, taking reciprocals gives
\[
1
\leq
K_{\mathrm{BEH}}
\leq
K_{\mathrm{eff}}
\leq
K.
\]
Thus, $K_{\mathrm{BEH}}$ further reduces the effective state count according to behavioral redundancy. In particular, if all distinct states are treated as behaviorally unrelated,
$
S^{\mathrm{BEH}}=I,
$
then
\[
K_{\mathrm{BEH}}
=
\frac{1}{p^\top p}
=
\frac{1}{\sum_{k=1}^{K}p_k^2}
=
K_{\mathrm{eff}}.
\]
Therefore, the occupancy-based effective count $K_{\mathrm{eff}}$ is a special case of $K_{\mathrm{BEH}}$ when all states are behaviorally distinct. If, in addition, the occupancies are equal, $p_k=1/K$ for all $k$, then $K_{\mathrm{BEH}}=K$. On the other hand, if every state is behaviorally indistinguishable from every other state,
$
S^{\mathrm{BEH}}
=
\mathbf{1}\mathbf{1}^\top,
$
then
$
p^\top S^{\mathrm{BEH}}p
=
\left(\sum_{k=1}^{K}p_k\right)^2
=
1,
$
and therefore
$
K_{\mathrm{BEH}}=1.
$

A completely unsupported nominal state with $p_k=0$ does not increase the effective complexity. In addition, $K_{\mathrm{BEH}}$ is invariant to state labeling, as simultaneously permuting the hidden-state labels in $p$ and $S^{\mathrm{BEH}}$ leaves $p^\top S^{\mathrm{BEH}}p$ unchanged \citep{leinster2012}. If two states are behaviorally identical, merging them into a single state with combined posterior occupancy leaves $K_{\mathrm{BEH}}$ unchanged. Moreover, because $K_{\mathrm{BEH}}$ is continuous in $p$ and $S^{\mathrm{BEH}}$, nearly identical states produce only a small change in effective behavioral complexity \citep{leinster2012}.

$K_{\mathrm{BEH}}$ is not the selected model order, as it can be non-integer, but it measures the effective behavioral complexity represented within a fitted $K$-state HMM after accounting for both posterior occupancy and behavioral similarity. Instead, this motivates studying the effective-complexity curve
\[
K\mapsto K_{\mathrm{BEH}}(K;\alpha),
\]
as the model-selection problem concerns how $K_{\mathrm{BEH}}$ changes as the candidate nominal order $K$ increases. Therefore, for each candidate model order $K$ and each value of $\alpha$, the fitted HMM is mapped to its behavioral effective complexity through
\[
\widehat{\theta}_K
\rightarrow
p
\rightarrow
D^{\mathrm{em}}
\rightarrow
D^{\mathrm{BEH}}
\rightarrow
S^{\mathrm{BEH}}
\rightarrow
K_{\mathrm{BEH}}(K;\alpha).
\]
Section 3 develops the procedure used to convert the effective-complexity curve into $\widehat K$.

\section{Behavioral-Complexity Model-Order Selection}

\subsection{Effective-Complexity Curves}

BCMOS converts each effective-complexity curve $K \mapsto K_{\mathrm{BEH}}(K;\alpha)$ for each $\alpha$ into a single estimated order through:
\[
K_{\mathrm{BEH}}(K;\alpha)
\rightarrow
\kappa_{\mathrm{seg}}(\alpha)
\rightarrow
\text{stability across }\alpha
\rightarrow
\widehat K.
\]
For each value of $\alpha$, we define the effective-complexity curve that describes how the behavioral effective complexity of the fitted HMM changes as the candidate nominal order $K$ increases:
$
B_K(\alpha)
=
K_{\mathrm{BEH}}(K;\alpha).
$
For smaller values of $K$, adding nominal states is expected to introduce new latent structure that makes $K_{\mathrm{BEH}}$ increase fast as it captures them. Once all the latent structure has been represented, further increases in $K$ introduce states that are weakly occupied, behaviorally similar to existing states, or both, and as $K_{\mathrm{BEH}}$ accounts for all that, such additional nominal states contribute less to the effective-complexity curve than genuinely distinct and well-supported states. Furthermore, we use changes in the slope of the effective-complexity curve to identify a candidate breakpoint at which additional nominal states provide a smaller increase in behavioral complexity.

\subsection{Segmented-Regression Breakpoint}

We fit a continuous two-segment piecewise-linear regression to the effective-complexity curve for each fixed value of $\alpha$ \citep{muggeo2003}. The segmented regression model is
\[
B_K(\alpha)
=
a_\alpha
+
b_{1,\alpha}K
+
\delta_\alpha(K-\kappa)_+
+
\varepsilon_K,
\]
where
$
(x)_+=\max(x,0),
$
and $\kappa$ denotes a candidate breakpoint. The slope before the breakpoint is
$
s_{\mathrm{pre}}(\alpha)=b_{1,\alpha},
$
while the slope after is
$
s_{\mathrm{post}}(\alpha)=b_{1,\alpha}+\delta_\alpha.
$
Moreover, for each value of $\alpha$, we fit the segmented regression at every candidate $\kappa$, denoted by $\widehat B_K(\alpha;\kappa)$, and measure the fitting error by
\[
\operatorname{RMSE}(\kappa;\alpha)
=
\left[
\frac{1}{|\mathcal K|}
\sum_{K\in\mathcal K}
\left(
B_K(\alpha)
-
\widehat B_K(\alpha;\kappa)
\right)^2
\right]^{1/2},
\]
with the segmented-regression breakpoint being defined as
\[
\kappa_{\mathrm{seg}}(\alpha)
=
\arg\min_{\kappa}
\operatorname{RMSE}(\kappa;\alpha).
\]

\subsection{Stability Across $\alpha$}

We examine the breakpoint path
$
\alpha \mapsto \kappa_{\mathrm{seg}}(\alpha)
$
as the segmented-regression breakpoint can vary with $\alpha$. Let
\[
\mathcal A=\{\alpha_1,\ldots,\alpha_M\},
\qquad
\alpha_1<\cdots<\alpha_M,
\]
be the fixed grid, and let
$
P=\{\alpha_r,\ldots,\alpha_s\}
$
be a maximal contiguous plateau over which $\kappa_{\mathrm{seg}}(\alpha)$ remains constant.
Moreover,
$
|P|=s-r+1
$
is the length of a plateau, and because the $\alpha$ grid is equally spaced, maximizing the number of $\alpha$ values in a plateau is equivalent to selecting the widest stability region on the grid.

Let $\mathcal P$ denote the collection of maximal contiguous plateaus. We select the longest plateau, breaking ties by the earliest starting value $\alpha_{\mathrm{start}}(P)$. Equivalently,
\[
P^\star
=
\underset{P\in\mathcal P}{\operatorname{argmax}^{\mathrm{lex}}}
\left(
|P|,-\alpha_{\mathrm{start}}(P)
\right).
\]
If the selected plateau $P^\star$ has constant breakpoint $k^\star$, the final model-order estimate is
$
\widehat K=k^\star.
$

\begin{algorithm}[H]
\caption{Behavioral-Complexity Model-Order Selection}
\label{alg:model_order_selection}
\begin{algorithmic}
\STATE {\bfseries Input:} observations $Y_{1:T}$, candidate orders $\mathcal K$, behavioral-weight grid $\mathcal A$

\FOR{$K\in\mathcal K$}
    \STATE Fit the $K$-state Gaussian HMM using multiple EM restarts and retain the highest-likelihood fit $\widehat{\theta}_K$
    \STATE Compute posterior occupancies $p$
    \FOR{$\alpha\in\mathcal A$}
        \STATE Compute $D^{\mathrm{em}}$, $D^{\mathrm{BEH}}$, $S^{\mathrm{BEH}}$, and $K_{\mathrm{BEH}}(K;\alpha)$
    \ENDFOR
\ENDFOR

\FOR{$\alpha\in\mathcal A$}
    \STATE Form $B_K(\alpha)=K_{\mathrm{BEH}}(K;\alpha)$ over $K\in\mathcal K$
    \STATE Fit the segmented regression and obtain $\kappa_{\mathrm{seg}}(\alpha)$
\ENDFOR

\STATE Construct maximal contiguous plateaus of $\kappa_{\mathrm{seg}}(\alpha)$
\STATE Select the longest plateau; break ties by the earliest starting $\alpha$
\STATE {\bfseries Output:} $\widehat K$, the breakpoint associated with the selected plateau
\end{algorithmic}
\end{algorithm}

\section{Synthetic Experiments}

\subsection{Simulation Design}

To study the accuracy of our method and to test the new algorithm for estimating the number of states, the performance of the proposed approach is evaluated on Gaussian Hidden Markov Models with the known true number of states \(K_0 \in \{3, 8\}\). For \(K_0 = 3\), models with the number of states \(K \in \{1, \ldots, 15\}\) and the sequence length \(T = 2000\) are fitted, while for \(K_0 = 8\), models with the number of states \(K \in \{1, \ldots, 20\}\) and \(T = 10000\) are fitted using independently generated datasets using Monte Carlo simulation blocks.

Conditional on the hidden state \(Z_t = k\), the observation distribution is
$
Y_t \mid Z_t=k \sim \mathcal{N}(\mu_k,\sigma^2),
$
with the uniform initial hidden-state distribution
$
\pi_k=\frac{1}{K_0}, \qquad k=1,\ldots,K_0,
$
the equally spaced emission means
$
\mu_k=k, \qquad k=1,\ldots,K_0,
$
and the same value of standard deviation for all states
$
\sigma \in \{0.3,0.5,0.8\}.
$

Moreover, we use the four transition-matrix families considered by \citep{chen2024} to vary the temporal dependence of the hidden process. Let \(E\) be the \(K_0\times K_0\) matrix of ones, \(I\) the identity matrix, and the four transition regimes:
$
P_1=\frac{1}{K_0}E
$
is the matrix with uniformly distributed rows,
$
P_2=
\left(
0.8-\frac{0.2}{K_0-1}
\right)I
+
\frac{0.2}{K_0-1}E
$
and
$
P_3=
\left(
0.95-\frac{0.05}{K_0-1}
\right)I
+
\frac{0.05}{K_0-1}E
$
have diagonal probabilities \(0.8\) and \(0.95\), respectively, to represent increasingly persistent latent-state dynamics, and finally
$
P_4=
\frac{0.9}{K_0-1}E
-
\left(
\frac{0.9}{K_0-1}-0.1
\right)I
$
places probability \(0.1\) on the current state and distributes the remaining probability across the other states.

In total, \(1200\) datasets are generated for each true number of states \(K_0\) by generating \(100\) independent datasets for every \((P_j,\sigma)\), and \(2400\) datasets across the two Monte Carlo blocks. Furthermore, the Gaussian Hidden Markov Model is estimated using \(75\) independent EM restarts for each candidate order and for each independently generated dataset, and then the \(K\)-state model with the maximum observed-data log-likelihood is selected among the successful runs. Both selectors are computed on the same best-of-\(75\) fit at each candidate order in order to avoid potential issues coming from different outcomes of independent optimization runs. All calculations related to the Gaussian HMM inference were implemented from scratch in Python, following the standard forward-backward and Baum-Welch recursions \citep{baum1970,rabiner1989,cappe2005}, and no preexisting HMM estimator packages were used.

\subsection{Model-Order Recovery: Proposed Method versus BIC}

For every dataset generated according to Section 4.1, both the proposed model-order selection criterion and BIC \citep{schwarz1978} are computed. 
For a fitted $K$-state model with observed-data log-likelihood $\ell_K$ and $p_K$ free parameters,
$
\operatorname{BIC}(K)
=
-2\ell_K+p_K\log T,
$
and the selected order is
$
\widehat K_{\mathrm{BIC}}
=
\arg\min_{K\in\mathcal K}
\operatorname{BIC}(K).
$
The primary evaluation criterion is exact recovery since the generating order $K_0$ is known in the simulations. For each $(P_j,\sigma)$ condition and $R=100$ independent datasets, the estimated exact-recovery probability is
$
\widehat{\rho}
=
\frac{1}{R}
\sum_{r=1}^{R}
\mathbf{1}
\left\{
\widehat K^{(r)}=K_0
\right\},
$
the proportion of selections lying within one state of the true order is
\[
\widehat{\Pr}
\left(
\left|\widehat K-K_0\right|\leq 1
\right)
=
\frac{1}{R}
\sum_{r=1}^{R}
\mathbf{1}
\left\{
\left|\widehat K^{(r)}-K_0\right|\leq 1
\right\},
\]
the mean absolute error is
$
\operatorname{MAE}
=
\frac{1}{R}
\sum_{r=1}^{R}
\left|
\widehat K^{(r)}-K_0
\right|,
$
to compute how large the errors are on average, and the mean signed error is
$
\operatorname{Bias}
=
\frac{1}{R}
\sum_{r=1}^{R}
\left(
\widehat K^{(r)}-K_0
\right),
$
to identify whether the methodology tends to underselect or overselect.

Table 1 reports the aggregate performance across the 12 $(P_j,\sigma)$ conditions for each value of $K_0$.

\begin{table}[t]
  \caption{Model-order recovery performance of the proposed method and BIC across the two Monte Carlo experiments. Exact and Within 1 are reported as percentages.}
  \label{tab:model-order-recovery}
  \begin{center}
    \begin{small}
      \begin{sc}
        \begin{tabular}{llccc}
          \toprule
          $K_0$ & Method & Exact & Within 1 & MAE \\
          \midrule
          3 & BCMOS & 78.92\% & 79.58\% & 1.329 \\
          3 & BIC   & 66.67\% & 92.33\% & 0.410 \\
          \midrule
          8 & BCMOS & 17.17\% & 40.50\% & 2.561 \\
          8 & BIC   & 48.50\% & 61.25\% & 1.737 \\
          \bottomrule
        \end{tabular}
      \end{sc}
    \end{small}
  \end{center}
\end{table}

As seen from Table 1, in the case $K_0 = 3$, BCMOS provides an exact-recovery rate of $78.92\%$, while BIC provides $66.67\%$. With respect to the within-one recovery rate, BIC provides $92.33\%$, whereas the proposed method provides $79.58\%$. Moreover, BIC has a lower value of mean absolute error, $0.410$ compared to $1.329$ provided by the proposed method. Besides, for $K_0 = 8$, the proposed method has an exact-recovery rate of $17.17\%$, compared to $48.50\%$ obtained by BIC. The mean signed errors of both methods are negative, and their values are $-1.874$ in the case of the proposed method and $-1.712$ in the case of BIC, which points to the tendency to underselect the models in the higher-order situation. Therefore, even though the proposed method gives a better exact-recovery rate at $K_0 = 3$, at $K_0 = 8$ the model-order recovery performance decreases considerably.

\subsection{Locating the loss of the order information}

In order to explain the drop in recovery performance for $K_0 = 8$, we first examine whether the nominal-order information is lost in the computation of the behavioral-distance metric or in the summarization of the behavioral representation by $K_{\mathrm{BEH}}$.
Since
$
S^{\mathrm{BEH}}
=
\mathbf 1\mathbf 1^\top-D^{\mathrm{BEH}}
$
and $p^\top\mathbf 1=1$, the behavioral effective complexity can
equivalently be written as
\[
K_{\mathrm{BEH}}
=
\frac{1}
{1-p^\top D^{\mathrm{BEH}}p}
=
\frac{1}
{1-2\sum_{i<j}p_i p_j D^{\mathrm{BEH}}_{ij}}.
\]
In our framework, the entire behavioral geometry is used in $K_{\mathrm{BEH}}$ only via a single occupancy-weighted summary. To examine whether scalar compression removes information that is useful for identifying the nominal order, we perform a held-out diagnostic using three representations: the local behavior of $K_{\mathrm{BEH}}$, the same information augmented with summaries of the pairwise behavioral geometry, and the full pre-compression representation including occupancy and similarity information. For $K_0=8$, the correct localization rate increases from $15.34\%$ using only $K_{\mathrm{BEH}}$ to $53.23\%$ using the full representation before compression. The complete diagnostic construction and detailed results are reported in Appendix~\ref{app:compression_diagnostics}. Finally, we examine the information in $D^{\mathrm{BEH}}$ by asking whether an additional fitted state results in an unusually inexpensive behavioral merge in transitioning from $K_0$ to $K_0 + 1$. State-merging approaches have been used before to prune Hidden Markov Models by finding equivalent or redundant states \citep{bicego2001}, while pairwise merge costs have a long history in hierarchical clustering methods \citep{ward1963}.

For states $i$ and $j$, we define the occupancy-weighted merge distortion as
\[
M_{ij}
=
\frac{2p_ip_j}{p_i+p_j}
D^{\mathrm{BEH}}_{ij},
\]
and the normalized minimum merge distortion as
\[
M_{\mathrm{rel}}(K)
=
\frac{\min_{i<j}M_{ij}}
{\operatorname{median}_{i<j}M_{ij}}.
\]
The merge distortions are directly related to the scalar behavioral
complexity. Indeed,
\[
p^\top D^{\mathrm{BEH}}p
=
2\sum_{i<j}p_ip_jD^{\mathrm{BEH}}_{ij}
=
\sum_{i<j}(p_i+p_j)M_{ij},
\]
and therefore
\[
K_{\mathrm{BEH}}
=
\left[
1-\sum_{i<j}(p_i+p_j)M_{ij}
\right]^{-1}.
\]
Thus, $K_{\mathrm{BEH}}$ aggregates the pairwise merge distortions
into a single weighted quantity, whereas $M_{\mathrm{rel}}$ retains
information about whether a particular pair is unusually cheap to
merge relative to the remaining pairs.

\begin{proposition}[Scalar compression does not identify local reducibility]
\label{prop:scalar_compression}
The behavioral effective complexity does not, in general, determine
the local pairwise reducibility of the behavioral geometry. In
particular, there exist distinct $[0,1]$-valued pseudometric matrices
$D$ and $D'$ and the same occupancy vector $p$ such that
$
K_{\mathrm{BEH}}(p,D)
=
K_{\mathrm{BEH}}(p,D'),
$
while
$
M_{\mathrm{rel}}(p,D)
\neq
M_{\mathrm{rel}}(p,D').
$
Hence, equal behavioral effective complexity does not imply equal
local redundancy structure.
\end{proposition}

A constructive example is provided in Appendix~\ref{app:proofs}.

The corresponding boundary statistic is
\[
\begin{aligned}
J_M(K)
&=
\log\!\left(M_{\mathrm{rel}}(K-1)+\varepsilon\right)
-
\log\!\left(M_{\mathrm{rel}}(K)+\varepsilon\right),\\
\varepsilon
&=10^{-12},
\end{aligned}
\]
where a large positive value indicates that moving from $K-1$ to $K$ creates a substantially cheaper behavioral merge. This reducibility analysis is used only as a truth-aware diagnostic and is not proposed as a new model-order selector. For $K_0 = 8$, the modal location across $\alpha$ of the maximum $J_M(K)$ marks the true overfitting boundary $K_0 + 1 = 9$ in $55.83\%$ of the datasets, and the jump at the true boundary is positive for $81.94\%$ of the $\alpha$ values on average. The signal, however, becomes much stronger in persistent $P_2/P_3$ regimes, where the true boundary is identified in $91.0\%$, $89.5\%$, and $81.5\%$ of the datasets for $\sigma = 0.3$, $0.5$, and $0.8$, respectively. More importantly, it can also be observed in many of those simulations when the order proposed by the method is far from the true one. Among the $551$ datasets where the method suggests choosing $K \in \{3, 4, 5\}$ given $K_0 = 8$, the behavioral-distance-based diagnostic assigns the modal location across $\alpha$ of the maximum $J_M(K)$ to $K = 9$ in $40.29\%$ of these cases. In addition, the median normalized merge distortion is $0.825$ at the suggested order compared to $0.242$ at $K = 9$, which indicates that, in aggregate, the cheapest merge is substantially more pronounced at $K = 9$ than at the order chosen by the scalar criterion. Therefore, the diagnostics indicate that the information about nominal order is not completely lost in the process of $D^{\mathrm{BEH}}$ computation. Rather, the behavioral geometry in many cases contains information about the overfitting onset, especially in the regime of persistent dynamics, while some of it gets lost during the reduction to the single scalar $K_{\mathrm{BEH}}$. Thus, the compression seems to be an important source of estimation error for the current methodology, albeit not the only one in some less persistent and noisier settings.

\section{Application to Limit Order Book Data}

\subsection{Data and Experimental Setup}

We test the methodology on the Microsoft Corporation (MSFT) limit order book MBP-10 schema dataset from January 26, 2026, obtained from the Nasdaq TotalView--ITCH feed through Databento. A limit order book records the outstanding supply and demand at several price levels and changes as orders are submitted, cancelled, or executed \citep{krause2022two,jung2024attention}. For the data cleaning step, we removed locked or crossed quotes, records with missing price or size values, and observations with invalid or nonpositive best-level depth, and then ordered the observations using the exchange event timestamp. For the analysis step, we used the change in the best-level midprice
$
\Delta m_t=m_t-m_{t-1},
$
with
$
m_t=\frac{p_{0,t}^{b}+p_{0,t}^{a}}{2},
$
the best-level order-flow imbalance \(\mathrm{OFI}_{0,t}\), and the inverse depth
$
\mathrm{ID}_{0,t}=\frac{1}{q_{0,t}^{b}+q_{0,t}^{a}},
$
where \(p_{0,t}^{b}\) and \(p_{0,t}^{a}\) are the best bid and ask prices, and where \(q_{0,t}^{b}\) and \(q_{0,t}^{a}\) are the best-level bid and ask sizes. The midpoint gives a measure of the price from both the buying and selling sides. Moreover, the order-flow imbalance shows the changes in supply and demand at the best bid and ask queues, while the book depth shows the amount of liquidity available to absorb incoming orders \citep{bouchaud2009price,cont2011price,krause2022two}.

Finally, the observation vector used for fitting the HMM is

$
X_t=
\begin{pmatrix}
\Delta m_t\\
\mathrm{OFI}_{0,t}\\
\mathrm{ID}_{0,t}
\end{pmatrix}.
$
Before fitting the models, we standardized each feature to have a sample mean of zero and a sample variance of one. After the cleaning step, we kept the complete trading day with \(T=825{,}336\) event-time observations. Then, we fitted HMMs for candidate orders from \(K=1\) to \(K=30\). For every candidate order, we used 75 independent EM restarts and kept the successful fit with the largest final observed-data log-likelihood. Finally, we used these same fitted models to compute both the proposed behavioral criterion and BIC.

\subsection{Model-Order Selection}

\begin{figure}[H]
    \centering
    \includegraphics[width=\columnwidth]{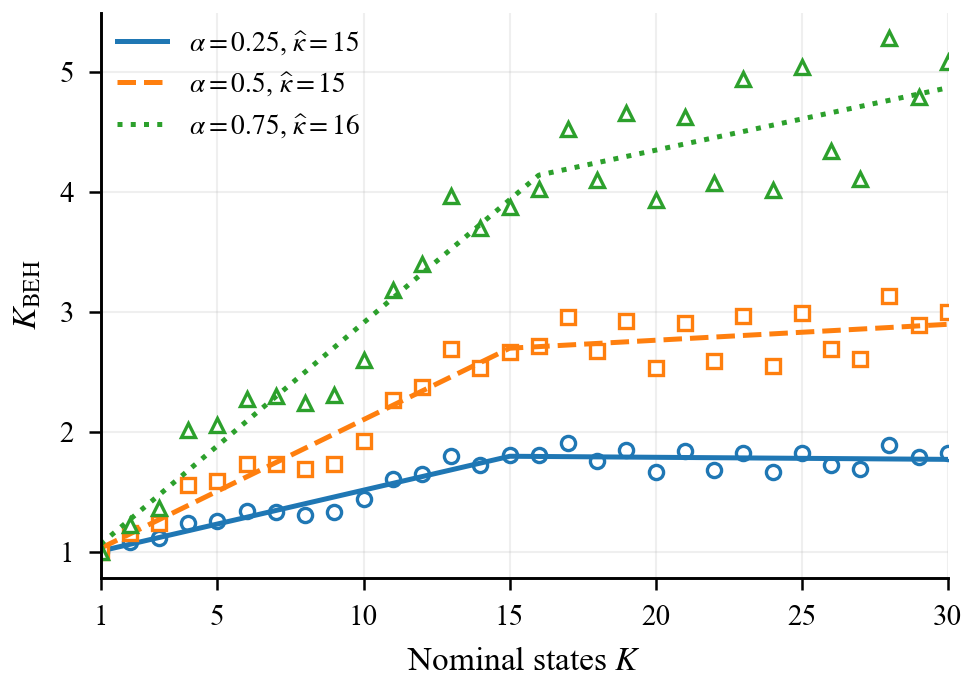}
    \caption{Continuous segmented-regression fits for $\alpha\in\{0.25,0.50,0.75\}$.}
    \label{fig:lob_segmented_fits}
\end{figure}

\begin{figure}[H]
    \centering
    \includegraphics[width=\columnwidth]{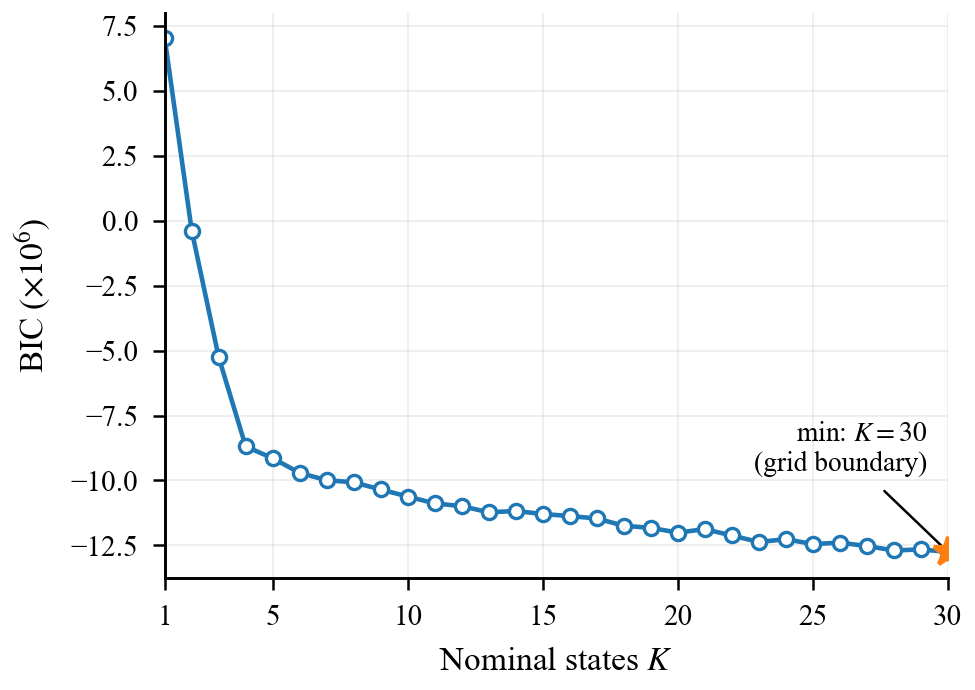}
    \caption{BIC across the candidate-order grid.}
    \label{fig:lob_bic}
\end{figure}

We used three alpha values to show the segmented-regression fits in Figure~\ref{fig:lob_segmented_fits}. The estimated breakpoint for alpha values 0.25 and 0.50 is \(\hat{k}=15\), while it is 16 for \(\alpha=0.75\). Also, the growth in behavioral complexity slowed down around \(K=15\) to \(K=16\), as the fitted slope decreases substantially in the three fits: from 0.0564 before the breakpoint to -0.0017 after it for \(\alpha=0.25\), from 0.1190 to 0.0133 for \(\alpha=0.50\), and from 0.2050 to 0.052 for \(\alpha=0.75\). Moreover, based on the longest stable breakpoint region, the estimated breakpoint occurs at \(\hat{K}=15\) with \(\alpha\in[0.05,0.55]\) over the whole grid \(\alpha\in\{0.05,0.10,\ldots,1.00\}\). On the other hand, based on Figure~\ref{fig:lob_bic}, BIC chooses the upper boundary of the search grid, \(K=30\), as the optimum.

This empirical study does not establish that either method estimated the true model order, as the true number of hidden states is unknown for the MSFT dataset. However, one main advantage of the proposed method over BIC is that the model-order selection directly accounts for the occupancies, emission profiles, and transition behavior of the fitted states, which also provides a behavioral interpretation of the selected model. More detailed results about the posterior occupancies, standardized emission means, and transition probabilities are shown in Appendix~\ref{app:selected_model}.

\section{Discussion and Conclusion}

Increasing the nominal number of hidden states can improve likelihood without introducing new latent structure, so we proposed, in this paper, a behavioral-complexity approach to select the model order of overparametrized Hidden Markov Models. The methodology evaluates whether additional fitted states contribute behavior that is supported by the observed data, by using the posterior state occupancy, and whether it is behaviorally distinct from existing states, by using the pairwise emission and transition similarities, and then combines them to construct the behavioral effective complexity \(K_{\mathrm{BEH}}\). Moreover, we fitted the segmented regression to the effective-complexity curve of \(K_{\mathrm{BEH}}\) across candidate orders \(K\) to identify the breakpoint where the growth in behavioral complexity begins to slow. Then, we studied the stability of the breakpoint across different values of the behavioral weight \(\alpha\) to finally obtain the nominal model-order estimate.

Based on the synthetic experiments in this paper, the proposed method achieves a higher exact-recovery rate than BIC for \(K_0=3\), while for the larger true order \(K_0=8\), the performance of the method deteriorates. This motivated us to examine where the information about the nominal order is being lost. The experiments showed that transforming the full behavioral-distance geometry into \(K_{\mathrm{BEH}}\) compresses a large collection of pairwise relationships into a single scalar quantity. At the same time, the onset of unusually cheap behavioral merges often occurs close to the true overfitting boundary even when the final scalar-based selector underestimates the order. Therefore, these results indicate that scalar compression is an important source of information loss in the methodology. Furthermore, the behavioral-distance signal is also weaker in some noisy and less persistent regimes, which shows that scalar compression is not the only source of error. Overall, even when the scalar selector does not recover the correct order, the underlying behavioral geometry still contains useful model-order information.

Moreover, we applied the methodology to a real MSFT limit order book dataset with 825,336 event-time observations, fitting candidate orders from \(K=1\) to \(30\). The behavioral criterion selects \(K=15\) as a parsimonious order where the growth in distinct effective behavioral complexity starts to slow substantially, while BIC selects \(K=30\), the upper boundary of the candidate-order grid. Another advantage of the methodology, besides estimating the number of hidden states, is that the selected \(K=15\) model can also be interpreted as a meaningful behavioral representation of the observed limit order book dynamics, since it contains heterogeneous state occupancies, emission profiles, and transition behavior.

The natural theoretical direction of the paper is to determine under which conditions the behavioral-complexity breakpoint converges to the true nominal order \(K_0\). Also, as the behavioral geometry contains information that is lost when transformed to the scalar \(K_{\mathrm{BEH}}\), it motivates the discovery of model-order selectors that preserve more of the pairwise behavioral-distance information. Finally, to determine whether the observed behavior is specific to high-frequency financial data, future work could be testing the methodology on non-financial sequential datasets.

\section*{Acknowledgments}

The authors acknowledge the computational resources provided by the Hyak high-performance computing system at the University of Washington. Ghaleb Khalil would like to thank Professor Sewoong Oh for the knowledge and perspective gained through his course and for his continued support. He is also deeply grateful to his family, parents, godfather, and friends for their constant encouragement, support, and belief in him throughout his academic journey.

\section*{Impact Statement}

This paper presents methodological work on model-order selection for Hidden Markov Models, with an empirical application to limit order book data. The proposed methodology is intended to improve the interpretation and selection of latent-state models rather than to provide a trading strategy or automated financial decision system. As with other statistical and machine-learning methods applied to financial data, downstream use may be subject to model risk and limitations arising from the assumptions of the fitted model. Beyond these general considerations, we do not identify specific societal consequences that require additional discussion.


\bibliography{example_paper}
\bibliographystyle{icml2026}

\newpage
\appendix
\onecolumn
\section{Detailed Simulation Results by Condition}
\label{app:simulation_results}

Table~2 reports the exact model-order recovery rate for each combination of transition regime \(P_j\) and noise level \(\sigma\) across the \(R=100\) independently generated datasets from the Section~4.2 simulation results, so that we can better understand how the performance changes across the simulation settings.

\begin{table}[H]
  \caption{Exact model-order recovery rates (\%) for BCMOS and BIC across the simulation conditions. Each entry is based on $R=100$ independently generated datasets.}
  \label{tab:condition_exact_recovery}
  \centering
  \begin{small}
    \begin{sc}
      \begin{tabular}{llcccccc}
        \toprule
        & & \multicolumn{2}{c}{$\sigma=0.3$}
          & \multicolumn{2}{c}{$\sigma=0.5$}
          & \multicolumn{2}{c}{$\sigma=0.8$} \\
        \cmidrule(lr){3-4}
        \cmidrule(lr){5-6}
        \cmidrule(lr){7-8}
        $K_0$ & Regime
        & BCMOS & BIC
        & BCMOS & BIC
        & BCMOS & BIC \\
        \midrule

        3 & $P_1$ & 99  & 100 & 43  & 0   & 4   & 0   \\
        3 & $P_2$ & 100 & 100 & 100 & 100 & 79  & 16  \\
        3 & $P_3$ & 100 & 100 & 100 & 100 & 100 & 100 \\
        3 & $P_4$ & 100 & 100 & 98  & 84  & 24  & 0   \\

        \midrule

        8 & $P_1$ & 0   & 46  & 1 & 0   & 5 & 0   \\
        8 & $P_2$ & 100 & 99  & 0 & 100 & 0 & 0   \\
        8 & $P_3$ & 91  & 86  & 0 & 100 & 0 & 100 \\
        8 & $P_4$ & 2   & 51  & 4 & 0   & 3 & 0   \\

        \bottomrule
      \end{tabular}
    \end{sc}
  \end{small}
\end{table}

Firstly, for \(K_0=3\), the proposed method performs well in a large part of the simulation grid. It recovers the true order in all datasets under \(P_2\) when \(\sigma=0.3\) and \(0.5\), under \(P_3\) for all three noise levels, and under \(P_4\) when \(\sigma=0.3\). However, when the noise increases, the recovery rate becomes much lower in some regimes, especially under \(P_1\) and \(P_4\). BIC also performs very well in several of the easier settings, but its performance also decreases strongly in some of the noisier cases.

Moreover, the results become more difficult for \(K_0=8\). At the lowest noise level, \(\sigma=0.3\), the proposed method still recovers the true order in \(100\%\) of the datasets under \(P_2\) and \(91\%\) under \(P_3\). However, this performance does not continue when the noise increases. For example, under both \(P_2\) and \(P_3\), the recovery rate of the proposed method becomes very low for \(\sigma=0.5\) and \(0.8\). In comparison, BIC performs better in some of these settings, such as \(P_2\) with \(\sigma=0.5\), and \(P_3\) with \(\sigma=0.5\) and \(0.8\), where it recovers the true order in all datasets. On the other hand, both methods perform poorly in several settings under \(P_1\) and \(P_4\).

As a conclusion of this study, model-order recovery depends strongly on both the temporal dependence of the hidden-state process and the separation of the emission distributions, which motivates us to look more closely in Section~4.3 at the behavioral information before the final scalar criterion and study where the information about the true order starts to disappear.

\subsection{Additional Diagnostics for Scalar Compression}
\label{app:compression_diagnostics}

The pairwise-geometry representation additionally includes the mean and variance of pairwise similarity, the maximum similarity, the contribution of the strongest pair, the concentration of the pairwise similarity contributions, and their effective pair count. The full representation further includes $K_{\mathrm{eff}}$, occupancy concentration, the total similarity correction, and occupancy entropy. For each dataset, $\alpha$, and admissible candidate order $K$, local features are constructed from $K-1$, $K$, and $K+1$, and a standardized logistic regression is trained using five-fold cross-validation with entire datasets assigned to the folds to predict whether $K=K_0$. For each held-out dataset-$\alpha$ case, the candidate order with the largest predicted probability is taken as the localized order; this diagnostic is used only to measure the amount of order-relevant information retained by each representation and is not proposed as an alternative model-order selector.

More specifically, define the occupancy concentration and total similarity correction as
\[
C_{\mathrm{occ}}
=
\sum_{i=1}^{K} p_i^2,
\qquad
C_{\mathrm{sim}}
=
2\sum_{i<j} p_i p_j S^{\mathrm{BEH}}_{ij},
\]
so that
\[
K_{\mathrm{eff}}
=
\frac{1}{C_{\mathrm{occ}}},
\qquad
K_{\mathrm{BEH}}
=
\frac{1}{C_{\mathrm{occ}}+C_{\mathrm{sim}}}.
\]
For each pair $i<j$, let
$
w_{ij}
=
2p_i p_j,
\qquad
c_{ij}
=
w_{ij} S^{\mathrm{BEH}}_{ij}.
$
The occupancy-weighted mean and variance of pairwise similarity are
\[
\overline S
=
\sum_{i<j}
\frac{w_{ij}}
{\sum_{r<s} w_{rs}}
S^{\mathrm{BEH}}_{ij},
\qquad
V_S
=
\sum_{i<j}
\frac{w_{ij}}
{\sum_{r<s} w_{rs}}
\left(
S^{\mathrm{BEH}}_{ij}
-
\overline S
\right)^2.
\]

The maximum pairwise similarity is
$
S_{\max}
=
\max_{i<j}
S^{\mathrm{BEH}}_{ij}.
$
When $C_{\mathrm{sim}}>0$, define the normalized pairwise similarity contributions as
$
q_{ij}
=
\frac{c_{ij}}
{C_{\mathrm{sim}}},
$
from which the contribution of the strongest pair, the concentration of the pairwise similarity contributions, and their effective pair count are
\[
q_{\max}
=
\max_{i<j} q_{ij},
\qquad
H_{\mathrm{pair}}
=
\sum_{i<j} q_{ij}^2,
\qquad
N_{\mathrm{pair}}
=
\frac{1}
{H_{\mathrm{pair}}}.
\]
The occupancy entropy is
$
H_{\mathrm{occ}}
=
-\sum_{i:p_i>0} p_i \log p_i.
$
For every scalar summary $x_K$, five local features are constructed:
\[
x_K,
\qquad
\Delta^- x_K
=
x_K-x_{K-1},
\qquad
\Delta^+ x_K
=
x_{K+1}-x_K,
\]
\[
R_x(K)
=
\Delta^- x_K
-
\Delta^+ x_K,
\qquad
R_x^{\mathrm{rel}}(K)
=
\frac{
\Delta^- x_K-\Delta^+ x_K
}{
|\Delta^- x_K|+10^{-12}
}.
\]
The $K_{\mathrm{BEH}}$-only representation consists of these five local features applied only to $K_{\mathrm{BEH}}$. The pairwise-geometry representation additionally applies them to $\overline S$, $V_S$, $S_{\max}$, $q_{\max}$, $H_{\mathrm{pair}}$, and $N_{\mathrm{pair}}$, while the full representation applies them to all of these quantities together with $K_{\mathrm{eff}}$, $C_{\mathrm{occ}}$, $C_{\mathrm{sim}}$, and $H_{\mathrm{occ}}$. The diagnostic is evaluated on the grid $\alpha\in\{0.05,0.10,\ldots,1.00\}$, and the two candidate orders at each end of the fitted-order grid are excluded from the local analysis. The logistic-regression diagnostic uses standardized predictors, $C=1$, balanced class weights, the \texttt{lbfgs} solver, a maximum of $4000$ iterations, and a fixed random seed of $3701$. The five folds are assigned at the dataset level and balanced within each $(K_0,P_j,\sigma)$ condition, so that all candidate orders and $\alpha$ values from a given simulated dataset remain in the same fold. The candidate order $K$, the true order $K_0$, $\alpha$, transition regime, noise level, sample size, and dataset identity are not included as predictors.

For $K_0 = 8$, the held-out diagnostic that uses only $K_{\mathrm{BEH}}$ correctly identifies the true order in $15.34\%$ of dataset-$\alpha$ cases, but adding the information about the pairwise geometry increases the correct localization rate to $45.95\%$, and using the full representation before compression increases it even further to $53.23\%$. The gap is especially large for the transition regimes $P_2$ and $P_3$, where the correct localization rate increases from $1.46\%$ using only $K_{\mathrm{BEH}}$ to $68.87\%$ using the richer representation.

\section{Stability of the MSFT Breakpoint Across \texorpdfstring{$\alpha$}{alpha}}
\label{app:alpha_stability}

\begin{figure}[H]
    \centering
    \includegraphics[width=0.62\textwidth]{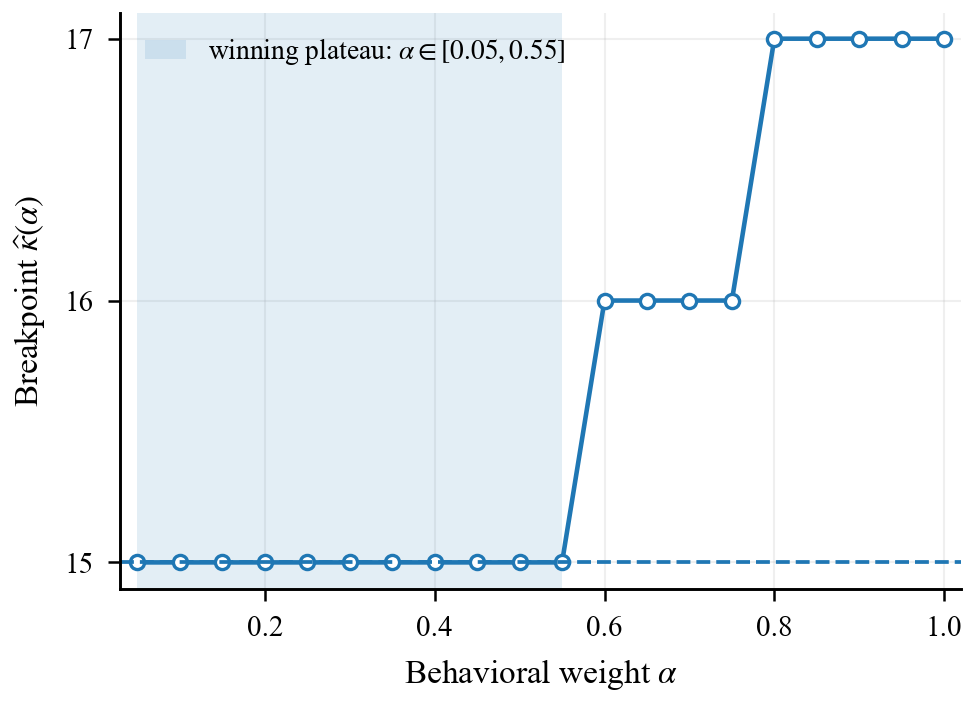}
    \caption{Estimated segmented-regression breakpoint across the complete behavioral-weight grid for the MSFT limit order book experiment. The shaded region denotes the winning plateau, where \(\widehat{\kappa}_{\mathrm{seg}}(\alpha)=15\) for \(\alpha\in[0.05,0.55]\).}
    \label{fig:alpha_plateau_stability}
\end{figure}

Figure~\ref{fig:alpha_plateau_stability} shows the estimated breakpoint across the complete \(\alpha\) grid. The longest contiguous stability region corresponds to \(\widehat{\kappa}_{\mathrm{seg}}(\alpha)=15\) for \(\alpha\in[0.05,0.55]\), covering more than half of the grid and therefore yielding the selected order \(K=15\). At larger values of \(\alpha\), which place greater weight on emission behavior relative to transition behavior, the estimated breakpoint moves to 16 and 17.

\section{Selected \texorpdfstring{$K=15$}{K=15} Model Diagnostics}
\label{app:selected_model}

The interpretation of the selected Gaussian HMM with \(K=15\), applied to the Microsoft (MSFT) limit order book dataset, is based on posterior state occupancies, standardized emission means, and transition dynamics. Important interpretations are: a positive inverse-depth mean implies lower-than-average best-level depth, and a negative inverse-depth mean implies greater-than-average depth; for the directional states, negative midprice changes are linked to negative Order Flow Imbalance (OFI) and positive midprice changes are linked to positive OFI; and similar emission means do not imply the same temporal dynamics. Table~\ref{tab:selected_state_interpretation} includes the main features of all 15 states.

\begin{figure}[H]
    \centering

    \begin{minipage}[t]{0.48\textwidth}
        \centering
        \includegraphics[width=\linewidth]{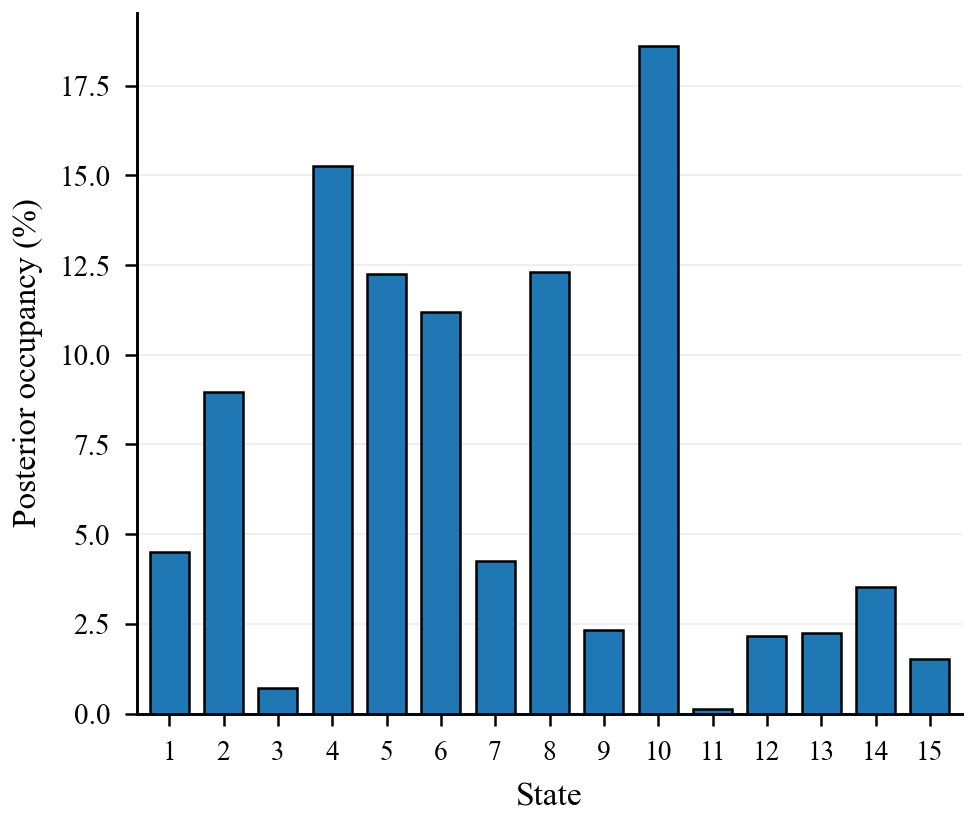}

        \small (a) Posterior state occupancy
    \end{minipage}
    \hfill
    \begin{minipage}[t]{0.48\textwidth}
        \centering
        \includegraphics[width=0.78\linewidth]{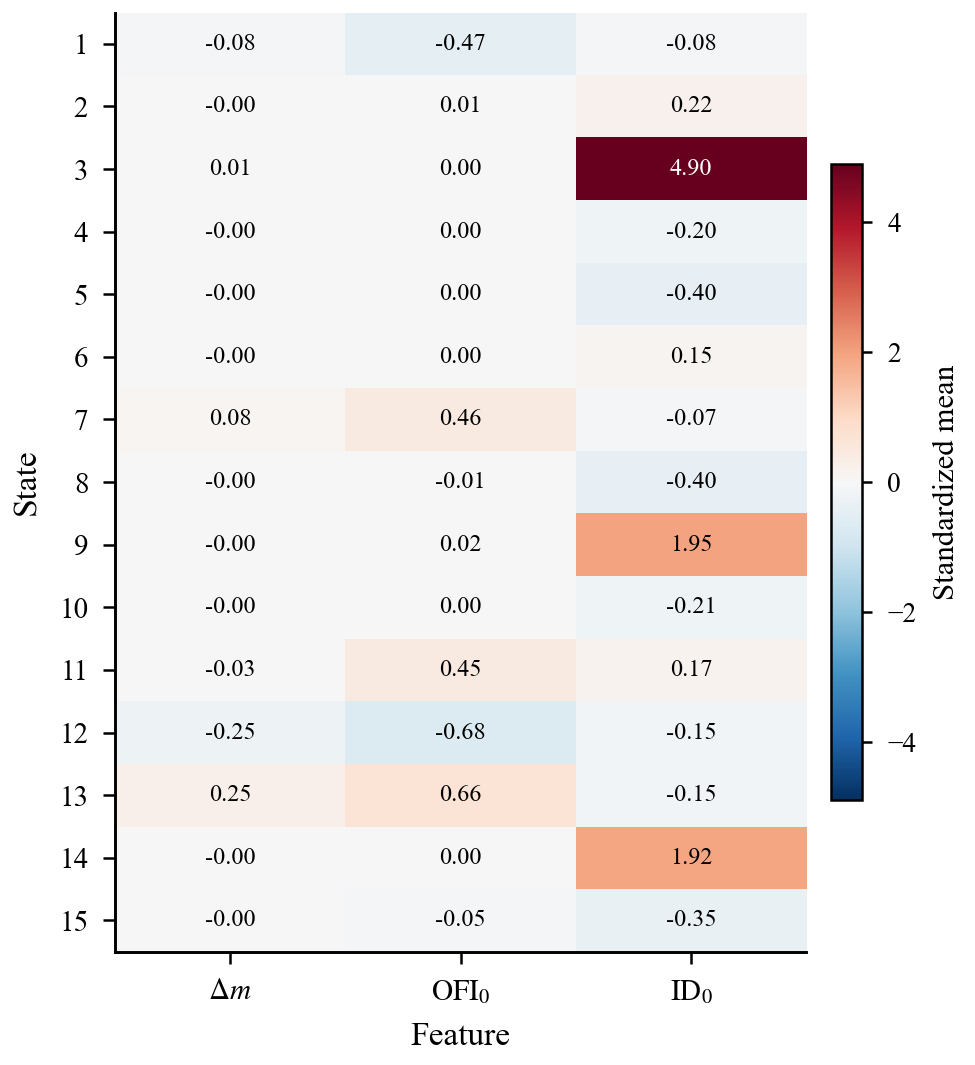}

        \small (b) Standardized emission means
    \end{minipage}

    \vspace{0.5em}

    \begin{minipage}[t]{0.55\textwidth}
        \centering
        \includegraphics[width=\linewidth]{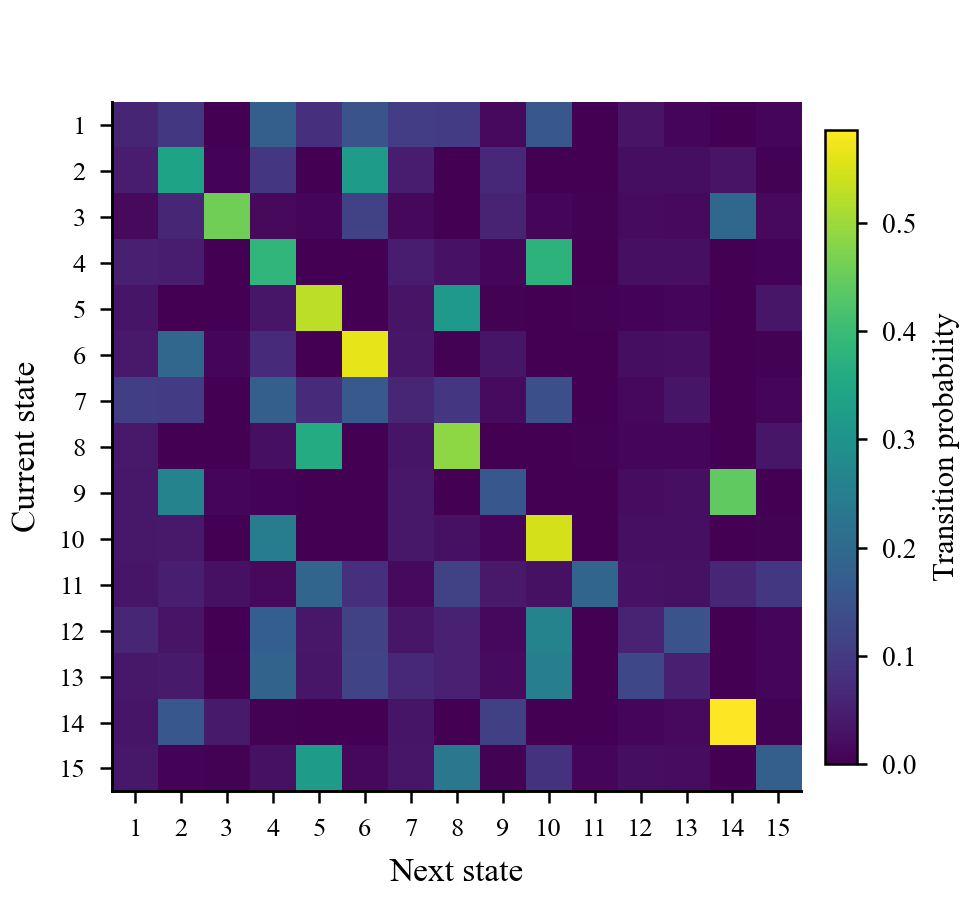}

        \small (c) Estimated transition matrix
    \end{minipage}

    \caption{
    Diagnostics for the selected \(K=15\) Gaussian HMM fitted to the
    MSFT limit order book data: (a) posterior state occupancies,
    (b) standardized emission means, and (c) estimated transition matrix.
    }
    \label{fig:selected_model_diagnostics}
\end{figure}

Figure~\ref{fig:selected_model_diagnostics}(a) shows that States 10, 4, 8, 5, and 6 are the most frequently occupied, accounting together for about 69.64\% of the posterior mass, with only States 3 and 11 having occupancies less than 1\%. State 3 is the most extreme thin-book state, with a standardized inverse-depth mean of 4.899 and an occupancy of 0.72\%, while State 11 has an occupancy of only 0.14\%. Thus, the state occupancies are unevenly distributed across the 15 states, but the selected model is not driven by states with very low posterior probability.

Figure~\ref{fig:selected_model_diagnostics}(b) suggests that States 1 and 7 form an approximately symmetric, moderately directional pair, whereas States 12 and 13 form a strong directional pair. Also, States 1, 7, 12, and 13 have low self-transition probabilities, meaning that these directional pairs are transient. States 4, 5, 8, 10, and 15 have negative inverse-depth means, meaning deeper-than-average best-level liquidity, while States 2, 3, 6, 9, and 14 have positive inverse-depth means, implying shallower-than-average books.

Figure~\ref{fig:selected_model_diagnostics}(c) shows that States 4 and 10 have very similar emission means and have strong bidirectional transitions between each other. State 10 is also more persistent than State 4, with \(A_{10,10}=0.550\) compared with \(A_{4,4}=0.387\). Similar pairs of states are (5, 8) and (2, 6), while State 9 strongly transitions to State 14. Lastly, directional states tend to make transitions to more frequently occupied near-neutral liquidity states; this can be seen from States 1 and 7 most strongly transitioning to State 4, and States 12 and 13 most strongly transitioning to State 10.

\begin{table}[H]
\centering
\small
\setlength{\tabcolsep}{4pt}
\caption{State-level interpretation of the selected \(K=15\) MSFT Gaussian HMM. Emission means are reported on the standardized feature scale. The final transition column reports the strongest non-self transition from each state.}
\label{tab:selected_state_interpretation}
\begin{tabular}{c r r r r r c p{4.3cm}}
\toprule
State &
Occ. (\%) &
\(\mu_{\Delta m}\) &
\(\mu_{\mathrm{OFI}}\) &
\(\mu_{\mathrm{ID}}\) &
\(A_{kk}\) &
Strongest transition &
Interpretation \\
\midrule
1 & 4.50 & -0.082 & -0.469 & -0.081 & 0.061 &
\(1\to4\) (0.178) &
Transient negative price/flow state with slightly greater depth \\

2 & 8.96 & 0.000 & 0.008 & 0.224 & 0.341 &
\(2\to6\) (0.322) &
Near-neutral state with moderately lower depth \\

3 & 0.72 & 0.013 & 0.000 & 4.899 & 0.459 &
\(3\to14\) (0.196) &
Rare extreme thin-book state \\

4 & 15.26 & 0.000 & 0.004 & -0.203 & 0.387 &
\(4\to10\) (0.379) &
High-occupancy deeper-book near-neutral state \\

5 & 12.26 & 0.000 & 0.003 & -0.396 & 0.527 &
\(5\to8\) (0.317) &
Persistent deep-book near-neutral state \\

6 & 11.20 & 0.000 & 0.003 & 0.146 & 0.564 &
\(6\to2\) (0.193) &
Persistent mildly shallow near-neutral state \\

7 & 4.24 & 0.082 & 0.458 & -0.069 & 0.064 &
\(7\to4\) (0.177) &
Transient positive price/flow state with slightly greater depth \\

8 & 12.31 & 0.000 & -0.013 & -0.398 & 0.486 &
\(8\to5\) (0.361) &
Deep-book near-neutral state with slightly negative OFI \\

9 & 2.34 & 0.000 & 0.017 & 1.953 & 0.160 &
\(9\to14\) (0.444) &
Transient thin-book near-neutral state \\

10 & 18.61 & 0.000 & 0.003 & -0.205 & 0.550 &
\(10\to4\) (0.246) &
Dominant persistent deeper-book near-neutral state \\

11 & 0.14 & -0.033 & 0.451 & 0.166 & 0.192 &
\(11\to5\) (0.191) &
Very rare atypical state \\

12 & 2.16 & -0.250 & -0.682 & -0.152 & 0.059 &
\(12\to10\) (0.261) &
Strong transient negative directional state \\

13 & 2.25 & 0.252 & 0.661 & -0.147 & 0.052 &
\(13\to10\) (0.250) &
Strong transient positive directional state \\

14 & 3.53 & 0.000 & 0.003 & 1.918 & 0.586 &
\(14\to2\) (0.160) &
Persistent thin-book near-neutral state \\

15 & 1.52 & -0.002 & -0.048 & -0.355 & 0.178 &
\(15\to5\) (0.325) &
Transient deep-book state with mildly negative OFI \\
\bottomrule
\end{tabular}
\end{table}

\section{Proofs}
\label{app:proofs}

\begin{proof}[Proof of Proposition~\ref{prop:behavioral_distance}]
Let \(D,D'\in\mathcal D_K\). For any pair of transition
distributions \(A_i,A_j\) and any coupling
\(\Gamma\in\Pi(A_i,A_j)\),
\[
\begin{aligned}
\left|
\sum_{\ell,m}\Gamma_{\ell m}D_{\ell m}
-
\sum_{\ell,m}\Gamma_{\ell m}D'_{\ell m}
\right|
&=
\left|
\sum_{\ell,m}\Gamma_{\ell m}
\left(D_{\ell m}-D'_{\ell m}\right)
\right| \\
&\leq
\|D-D'\|_\infty
\sum_{\ell,m}\Gamma_{\ell m}.
\end{aligned}
\]

Since \(A_i\) and \(A_j\) are probability vectors, every admissible
coupling has total mass one,
\[
\Gamma\in\Pi(A_i,A_j)
\quad\Longrightarrow\quad
\sum_{\ell,m}\Gamma_{\ell m}=1
\quad\Longrightarrow\quad
\left|
\sum_{\ell,m}\Gamma_{\ell m}D_{\ell m}
-
\sum_{\ell,m}\Gamma_{\ell m}D'_{\ell m}
\right|
\leq
\|D-D'\|_\infty.
\]

Let \(\Gamma'\) be an optimal coupling for
\(W_{D'}(A_i,A_j)\). Using the same coupling in the transport
problem with cost \(D\),
\[
\begin{aligned}
W_D(A_i,A_j)
&\leq
\sum_{\ell,m}\Gamma'_{\ell m}D_{\ell m} \\
&\leq
\sum_{\ell,m}\Gamma'_{\ell m}D'_{\ell m}
+
\|D-D'\|_\infty \\
&=
W_{D'}(A_i,A_j)+\|D-D'\|_\infty,
\end{aligned}
\]
and therefore
\[
W_D(A_i,A_j)-W_{D'}(A_i,A_j)
\leq
\|D-D'\|_\infty.
\]
Interchanging \(D\) and \(D'\) gives
\[
\begin{aligned}
W_{D'}(A_i,A_j)-W_D(A_i,A_j)
&\leq
\|D-D'\|_\infty, \\
\left|
W_D(A_i,A_j)-W_{D'}(A_i,A_j)
\right|
&\leq
\|D-D'\|_\infty.
\end{aligned}
\]

By the definition of \(\mathcal T_\alpha\),
\[
\begin{aligned}
\left|
\mathcal T_\alpha(D)_{ij}
-
\mathcal T_\alpha(D')_{ij}
\right|
&=
(1-\alpha)
\left|
W_D(A_i,A_j)-W_{D'}(A_i,A_j)
\right| \\
&\leq
(1-\alpha)\|D-D'\|_\infty.
\end{aligned}
\]
Taking the maximum over \(i,j\),
\[
\|\mathcal T_\alpha(D)-\mathcal T_\alpha(D')\|_\infty
\leq
(1-\alpha)\|D-D'\|_\infty.
\]

The Hellinger distance is a metric on probability distributions
\citep{gibbs2002}. Therefore,
\[
D^{\mathrm{em}}_{ij}
=
H(f_i,f_j)
\]
is a pseudometric on the hidden-state indices, since two distinct states may have identical emission distributions. Moreover,
\[
D\in\mathcal D_K
\quad\Longrightarrow\quad
D \text{ is a pseudometric on the finite state space}.
\]

By the standard properties of the discrete Wasserstein distance
\citep{peyre2019}, a pseudometric ground cost \(D\) induces a
pseudometric \(W_D\) on probability distributions:
\[
D\in\mathcal D_K
\quad\Longrightarrow\quad
W_D \text{ is a pseudometric}.
\]
Therefore, its pullback through the transition distributions,
\[
(i,j)\longmapsto W_D(A_i,A_j),
\]
is also a pseudometric on the hidden-state indices. Hence,
\[
\left.
\begin{array}{c}
D^{\mathrm{em}} \text{ is a pseudometric},\\[1mm]
(i,j)\mapsto W_D(A_i,A_j) \text{ is a pseudometric},\\[1mm]
\alpha\geq0,\qquad 1-\alpha\geq0
\end{array}
\right\}
\quad\Longrightarrow\quad
\alpha D^{\mathrm{em}}
+
(1-\alpha)W_D
\text{ is a pseudometric},
\]
and therefore
\[
\mathcal T_\alpha(D)
\text{ is a pseudometric}.
\]

Also,
\[
0\leq D^{\mathrm{em}}_{ij}\leq1,
\qquad
0\leq D_{\ell m}\leq1.
\]
Since every coupling has total mass one,
\[
0
\leq
W_D(A_i,A_j)
\leq
1.
\]
Therefore,
\[
0
\leq
\mathcal T_\alpha(D)_{ij}
=
\alpha D^{\mathrm{em}}_{ij}
+
(1-\alpha)W_D(A_i,A_j)
\leq
1,
\]
so that
\[
\mathcal T_\alpha(D)\in\mathcal D_K,
\qquad
\mathcal T_\alpha:\mathcal D_K\to\mathcal D_K.
\]

The set \(\mathcal D_K\) is defined by the symmetry, zero-diagonal,
boundedness, and triangle-inequality constraints of a pseudometric.
Each of these constraints is closed, so
\[
\mathcal D_K
\subset
\mathbb R^{K\times K}
\]
is closed. Since \(\mathbb R^{K\times K}\) is finite-dimensional and
therefore complete under the supremum norm,
\[
\mathcal D_K \text{ closed}
\quad\Longrightarrow\quad
(\mathcal D_K,\|\cdot\|_\infty)
\text{ is complete}.
\]

Combining the previous results, the Banach fixed-point theorem
\citep{kreyszig1978} applies:
\[
\left.
\begin{array}{c}
(\mathcal D_K,\|\cdot\|_\infty)
\text{ is complete},\\[1mm]
\mathcal T_\alpha:\mathcal D_K\to\mathcal D_K,\\[1mm]
\|\mathcal T_\alpha(D)-\mathcal T_\alpha(D')\|_\infty
\leq
(1-\alpha)\|D-D'\|_\infty,\\[1mm]
0\leq1-\alpha<1
\end{array}
\right\}
\quad\Longrightarrow\quad
\exists!\,
D^{\mathrm{BEH}}\in\mathcal D_K
\]
such that
\[
\mathcal T_\alpha(D^{\mathrm{BEH}})
=
D^{\mathrm{BEH}}.
\]
Moreover, for every \(D^{(0)}\in\mathcal D_K\),
\[
D^{(r+1)}
=
\mathcal T_\alpha(D^{(r)})
\quad\Longrightarrow\quad
D^{(r)}
\longrightarrow
D^{\mathrm{BEH}}.
\]
Finally, the standard contraction bound from the Banach fixed-point
theorem gives
\[
\|D^{(r)}-D^{\mathrm{BEH}}\|_\infty
\leq
(1-\alpha)^r
\|D^{(0)}-D^{\mathrm{BEH}}\|_\infty.
\]

Therefore, the behavioral distance has a unique fixed point and the fixed-point
iteration converges to it geometrically.
\end{proof}

\begin{proof}[Proof of Proposition~\ref{prop:scalar_compression}]
Let
\[
p=
\begin{pmatrix}
1/3\\
1/3\\
1/3
\end{pmatrix},
\]
and consider
\[
D=
\begin{pmatrix}
0 & 0.5 & 0.5\\
0.5 & 0 & 0.5\\
0.5 & 0.5 & 0
\end{pmatrix},
\qquad
D'=
\begin{pmatrix}
0 & 0.2 & 0.65\\
0.2 & 0 & 0.65\\
0.65 & 0.65 & 0
\end{pmatrix}.
\]
Both matrices are symmetric with zero diagonal and entries in
\([0,1]\), and their off-diagonal entries satisfy the triangle
inequalities; therefore, both are \([0,1]\)-valued pseudometrics.
Moreover,
\[
p^\top Dp
=
p^\top D'p
=
\frac{1}{3}.
\]
Since
\[
K_{\mathrm{BEH}}(p,D)
=
\frac{1}{1-p^\top Dp},
\]
it follows that
\[
K_{\mathrm{BEH}}(p,D)
=
K_{\mathrm{BEH}}(p,D')
=
\frac{3}{2}.
\]

For uniform occupancy,
\[
M_{ij}
=
\frac{1}{3}D_{ij}.
\]
For \(D\), all three off-diagonal merge distortions are equal, so
\[
M_{\mathrm{rel}}(p,D)=1.
\]
For \(D'\), the off-diagonal distances are
\(0.2,0.65,0.65\), and therefore
\[
M_{\mathrm{rel}}(p,D')
=
\frac{0.2}{0.65}
=
\frac{4}{13}.
\]
Therefore,
\[
M_{\mathrm{rel}}(p,D)
\neq
M_{\mathrm{rel}}(p,D'),
\]
despite the two geometries having exactly the same
\(K_{\mathrm{BEH}}\).
\end{proof}

\end{document}